\newcommand{\CLASSINPUTinnersidemargin}{56pt}
\newcommand{\CLASSINPUToutersidemargin}{56pt}
\newcommand{\CLASSINPUTtoptextmargin}{61pt}
\newcommand{\CLASSINPUTbottomtextmargin}{56pt}

\documentclass[letterpaper,10pt,conference]{IEEEtran}
\IEEEoverridecommandlockouts
\renewcommand{\IEEEtitletopspaceextra}{11pt}

\usepackage[utf8]{inputenc}
\usepackage[T1]{fontenc}
\usepackage{amsmath,amssymb,mathtools}
\usepackage{graphicx}
\usepackage{booktabs}
\usepackage{multirow}
\usepackage{url}
\usepackage{xcolor}
\usepackage{microtype}
\usepackage{cite}
\usepackage{caption}
\usepackage{flushend}
\usepackage[hidelinks]{hyperref}
\hypersetup{pdftitle={Keypose or Dense? A Controlled Comparison of Action Representations for Stage-Structured Manipulation},
  pdfauthor={Haizhou Ge, Mengshuo Wang, Zhixing Chen, Lei Han, Chen Wang, Lu Shi, Guyue Zhou, Ruqi Huang}, pdfsubject={}, pdfkeywords={}, pdfcreator={}}

\newcommand{\slotval}[1]{a_{#1}}
\newcommand{\stageend}[1]{e_{#1}}
\newcommand{\stageof}[1]{s(#1)}

\newcommand{\MainScenes}{300}
\newcommand{\HorizonStackPyramid}{400}
\newcommand{\HorizonLiftPegUpright}{350}
\newcommand{\StackPyramidTwoHundredDelta}{$-$1.8}
\newcommand{\StackPyramidTwoHundredCI}{[$-$8.0, +4.6]}
\newcommand{\MainScenesScaleLarge}{290}
\newcommand{\MainScenesScaleSmall}{287}
\newcommand{\MainScenesPosition}{245}
\newcommand{\FollowUpScenesScaleLarge}{293}
\newcommand{\DefaultExecDeltaID}{+6.4}
\newcommand{\DefaultExecDeltaLevelThree}{+39.0}
\newcommand{\DefaultExecDeltaScaleLarge}{+16.2}
\newcommand{\FreshScenesScaleSmall}{289}
\newcommand{\FreshScenesPosition}{251}
\newcommand{\DefaultExecDeltaScaleSmall}{+9.3}
\newcommand{\DefaultExecDeltaPosition}{+12.4}
\newcommand{\SelectedMinusDefaultDenseScaleSmall}{$-$6.2}

\newcommand{\ConditionsOtherScenes}{200}
\newcommand{\SyncSwapTuneBefore}{39}
\newcommand{\SyncSwapTuneAfter}{1.75}

\newcommand{\RealTrials}{20}
\newcommand{\RealLimit}{30}
\newcommand{\RealDenseSuccess}{17}
\newcommand{\RealKeyposeDefaultSuccess}{2}
\newcommand{\RealKeyposeAdjustedSuccess}{11}

\title{Keypose or Dense? A Controlled Comparison of Action Representations for Stage-Structured Manipulation}

\author{Haizhou Ge$^{1}$, Mengshuo Wang$^{1}$, Zhixing Chen$^{1}$, Lei Han$^{2}$,\\
  Chen Wang$^{1}$, Lu Shi$^{1\dag}$, Guyue Zhou$^{1\dag}$, Ruqi Huang$^{1\dag}$%
  \thanks{\hangindent=1em\hangafter=1\raggedright
    $^{1}$Tsinghua University. $^{2}$DISCOVER Robotics. $^{\dag}$Corresponding authors.\newline
    \mbox{\{ghz23, wangms25, chenzx24\}@mails.tsinghua.edu.cn},
    \mbox{\{wangchen, zhouguyue\}@air.tsinghua.edu.cn},
    \mbox{lus@nvidia.com},
    \mbox{ruqihuang@sz.tsinghua.edu.cn},
    \mbox{leo.han@discover-robotics.com}}}

\begin{document}
\bstctlcite{IEEEexample:BSTcontrol}

\maketitle

\begin{abstract}
\looseness=-1 
Most imitation policies that predict chunks of future actions use the dense action representation, in which the chunk holds one control target per step.
In stage-structured manipulation, where a task can be divided into stages with definable endpoints, the keypose action representation instead sets each step of the chunk to the endpoint of the stage that step belongs to, and a speed-capped servo supplies the motion.
We compare the two in several simulated manipulation tasks and on a real robot, with the same chunk length and matched demonstrations, observations, policy architectures and update budgets; in simulation, each one's execution rule is selected on tuning scenes.
When the simulated demonstrations are slowed, paused and rerouted, dense success falls sharply while keypose stays roughly level, across three policy architectures and with endpoints extracted from the recorded demonstrations; keypose also leads under object-size and position shift.
With clean demonstrations, keypose leads only when they are scarce and is otherwise close to or behind dense across the simulated tasks, provided that the demonstrations and execution rule suit its executor; when they do not, it falls far below dense.
The same dependence on the executor appeared on a real robot learning to open a drawer from human demonstrations: dense succeeded more often, and keypose improved substantially only after its executor's rotation and forward speed were adjusted to the demonstrations.
These results suggest that keypose trades sensitivity to such variation in the demonstrations for sensitivity to execution that departs from them.

\end{abstract}

\begin{IEEEkeywords}
Imitation learning, learning from demonstration, deep learning in grasping and manipulation
\end{IEEEkeywords}

\section{Introduction}
\label{sec:intro}

An imitation policy that predicts chunks of actions must decide what each slot of the chunk holds.
Most visuomotor policies write one control target per step, as in Action Chunking with Transformers (ACT)~\cite{zhao2023act}, Diffusion Policy~\cite{chi2023diffusion} and $\pi_0$ and $\pi_{0.5}$~\cite{black2025pi0,black2025pi05}.
Another line of work predicts the pose at which the current stage of the task ends and leaves the motion to a planner or controller~\cite{shridhar2022perceiver,xian2023chaineddiffuser}.
Both are in use, and we ask when either should be preferred under matched training and explicit execution selection.

Existing studies establish the usefulness and limitations of waypoint, endpoint and hybrid policies under their respective supervision and execution designs.
In the paired comparisons of automatic waypoint extraction (AWE), the gain from waypoint relabeling shrinks as demonstrations increase~\cite{shi2023waypoint}; HYDRA reports near-zero success for variants that predict only goal poses~\cite{belkhale2023hydra}; SkiP finds its variant without dense refinement far below its dense base policy~\cite{dai2026skip}.
We therefore test a shared-grid comparison and quantify how its results depend on endpoint source, execution selection and stopping.

\begin{figure*}[t]
  \centering
  \includegraphics[width=\textwidth]{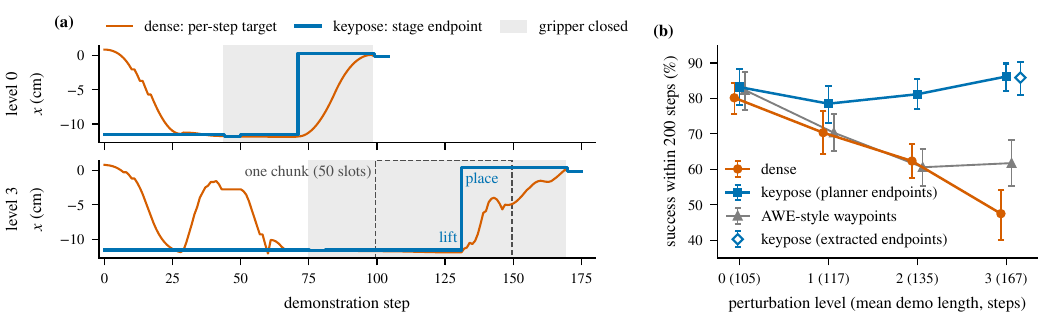}
  \caption{Keypose and dense action representations on a shared action-chunk grid.
  (a)~Dense and keypose action representations of two demonstrations of one StackCube scene along the world $x$ axis,
  without perturbation (top) and at perturbation level~3 (bottom). Each dense slot holds the controller's target
  for that step, so it copies the inserted detour and pause; each keypose slot holds the endpoint of the stage
  that slot belongs to. The dashed box is one 50-slot chunk; it spans the lift and place stages, so its
  keypose slots hold two endpoints. The gripper channel is the same in both. Both are trained with the same
  demonstrations, policy architecture and update budget and execute the first eight of 50 slots before querying
  again; the keypose command toward a stage endpoint is capped at the 95th percentile of the demonstrations' per-step motion.
  (b)~Success within 200 steps for ACT with default execution, trained on 100 demonstrations per level, on 300 test scenes, 4 training
  seeds per action representation and level, with 95\% intervals. As
  demonstrations are slowed, paused and rerouted, dense falls from 80.2\% to 47.5\%, while keypose with planner
  endpoints goes from 83.2\% to 86.2\%; with endpoints extracted from recorded commands and tool-center-point (TCP) poses by a
  frozen, task-configured procedure it reaches 85.8\% at level~3. Grey: AWE-style waypoints extracted from each demonstration instead of keypose. The result requires reliable endpoints, and
  demonstrations and execution rules compatible with the endpoint executor.}
  \label{fig:hero}
\end{figure*}

We place the two action representations on one action-chunk grid (Fig.~\ref{fig:hero}(a)): dense slots hold per-step control targets, while keypose slots hold the endpoint of the stage each slot belongs to and leave the path between endpoints to the executor.
Demonstrations, observations, policy architectures and update budgets are matched.
We found that running trained dense policies faster, without retraining, substantially raises their success on perturbed demonstrations, so each action representation's execution rule is selected on tuning scenes from candidates that include accelerated dense execution.
In simulation, when the demonstrations vary in how they move, being slowed, paused and rerouted without changing the task, dense success falls sharply while keypose stays roughly level (Fig.~\ref{fig:hero}(b)), also with endpoints extracted from the recorded demonstrations, but keypose falls far below dense when the demonstrations or execution rule do not suit its executor.
On a real robot learning from human demonstrations, this dependence showed again: dense succeeded more often, and keypose improved sharply once its executor was adjusted to move like the demonstrations, suggesting that keypose trades sensitivity to demonstration variation for sensitivity to execution that leaves the states they cover.

Our contributions are:
\begin{enumerate}
\setlength{\itemsep}{1pt}
\item A protocol for comparing action representations under matched training, with equal-budget execution selection and graded demonstration variation.
\item We show in simulation that keypose is far more robust than dense to demonstration variation, across policy architectures and with endpoints extracted from the demonstrations.
\item We show that keypose also leads when clean demonstrations are scarce and under object-size and position shift.
\item We identify the conditions keypose depends on, chiefly execution that follows the demonstrations, and confirm them on a real robot, where dense is more reliable.
\end{enumerate}

\section{Related Work}
\label{sec:related}

\textbf{Dense action chunks and goal poses.}
ACT, Diffusion Policy, $\pi_0$ and $\pi_{0.5}$ predict chunks of per-step actions, which FAST tokenizes for autoregressive vision-language-action (VLA) policies~\cite{pertsch2025fast}.
Why chunking helps is itself studied: Lazzati et al.~\cite{lazzati2026chunking} separate its benefit into non-Markovian expressiveness, reduced error propagation and implicit ensembling by delaying observations within one trained policy; Zeng et al.~\cite{zeng2026openloop} tie the benefit of long open-loop execution to non-Markovian demonstrations such as pauses; real-time chunking~\cite{black2025rtc} addresses the switch between chunks under inference delay.
Query timing and observation opportunities are thus competing explanations for differences between executions, and ours are not matched: accelerated dense execution queries more often than keypose.
A second family predicts the next gripper pose to reach and hands it to a motion planner~\cite{james2022c2farm,shridhar2022perceiver,goyal2023rvt,gervet2023act3d} or to a learned trajectory generator~\cite{xian2023chaineddiffuser,ma2024hdp,yu2024bikc}.
Keyposes are also labeled automatically from trajectories~\cite{lu2026keyposeexploration}, used as an interface for bimanual control~\cite{yang2025gripperkeypose} and added as auxiliary supervision in VLA fine-tuning~\cite{xu2026stake}.
Endpoint prediction is thus established; this paper contributes a controlled comparison, not a new policy family.
Our keypose differs in how the endpoint enters the policy: it supervises every future slot of a dense-length chunk with the endpoint of that slot's stage, shares the future gripper channel with dense, and is queried again every eight steps instead of being switched upon arrival.

\textbf{Comparisons and hybrids of action representations.}
AWE relabels demonstrations with waypoints that linearly reconstruct the trajectory, and in its paired RoboMimic results the gain shrinks as demonstrations increase.
Our AWE-style baseline extracts waypoints from positions only, keeps the dense future gripper channel and shares keypose's chunk length and execution without separate tuning, so it measures geometric waypoints extracted from perturbed demonstrations rather than AWE as published.
HYDRA switches between goal-pose and dense actions and, in its Appendix~D.4, discusses why goal-pose-only variants fail, including distribution shift and mismatch with the controller; we add quantitative interventions on demonstrations and execution rules under declared conditions.
SPHINX~\cite{sundaresan2024sphinx} and HoMeR~\cite{sundaresan2025homer} combine goal-pose actions for large motions with dense actions for precise ones, and SegDiff~\cite{cao2026segdiff} predicts the full trajectory segment to the next keypose; these designs address settings where endpoints alone do not suffice, which our comparison of stage-structured configurations does not cover.
Closest to our question, SkiP shows in its Table~4 that its variant without dense refinement, with the same $\pi_{0.5}$ policy architecture, falls far below its dense base, so removing dense refinement can be harmful.
Under a different chunk content, executor, stopping rule and demonstration set, our simulation results show that endpoint supervision can be more robust, while on our real robot dense succeeded more often.

\textbf{Irregular demonstrations and execution speed.}
Movement primitives separate the goal of a motion from its path and timing~\cite{ijspeert2002attractor,paraschos2013promp}.
ISR~\cite{yang2026isr} standardizes the timing of irregular human demonstrations, which contain speed changes and pauses, for dense policies.
SAIL~\cite{arachchige2025sail} and DemoSpeedup~\cite{guo2025demospeedup} make dense policies execute faster than their demonstrations.
DAgger~\cite{ross2011reduction} and DART~\cite{laskey2017dart} address coverage of the states a policy visits at deployment, which bears on executor paths that leave the demonstrated states.
We test stride acceleration of dense execution as a declared candidate family rather than reproducing SAIL or DemoSpeedup, and our ISR-style baseline is a controller-adapted variant with the original parameters, which tests that adaptation rather than strong dense preprocessing.

\section{Comparison Protocol}
\label{sec:protocol}

\subsection{Task and Observations}
The main task is ManiSkill3~\cite{tao2025maniskill3} StackCube: a Panda arm grasps a cube and places it on a second cube.
Policies observe two $128{\times}128$ RGB cameras and the proprioceptive state, and command end-effector delta poses at 20\,Hz.
Demonstrations come from the simulator's scripted motion-planning expert, whose six primitives (approach, descend, close, lift, place, open) define the stages.
Two further tasks are introduced in Section~\ref{sec:conditions}.

\subsection{Two Action Representations on One Chunk Grid}
\label{sec:representations}
An \emph{action representation} is what each slot of the action chunk holds, together with how executed slots become robot commands.
At step $t$ the policy outputs a chunk $(\slotval{t+k})_{k=0}^{Q-1}$ with $Q=50$, each slot an end-effector pose and a gripper command, and the first $n=8$ slots are executed before the next query.
Slot poses are relative to the tool-center-point (TCP) pose observed at the query, with translation offsets in world axes and rotations in the 6D rotation representation~\cite{zhou2019continuity}; slots past the end of a demonstration repeat its final TCP pose and last gripper command.
In the dense action representation, $\slotval{t+k}$ is the control target that the demonstration commanded at step $t+k$.
In the keypose action representation, the pose part of $\slotval{t+k}$ is $\stageend{\stageof{t+k}}$, the endpoint of the stage $\stageof{t+k}$ that contains step $t+k$, so a chunk that crosses a stage boundary holds two or more endpoints (Fig.~\ref{fig:hero}(a)); the gripper channel is the same as in dense.
A keypose slot is a stage endpoint, not the next step's target, so its execution needs a separate rule for how far to move per step (Section~\ref{sec:executors}).
That rule is part of the action representation, and the results therefore compare two configurations, not an effect of the chunk's content independent of execution.

Endpoints come from one of two sources.
\emph{Planner endpoints} are the target poses the scripted expert plans for each stage; the perturbations of Section~\ref{sec:perturbations} change only how the robot moves to these targets, so planner endpoints are the same at every perturbation level.
\emph{Extracted endpoints} come from a frozen, task-configured procedure that reads only the recorded commands and TCP poses of each demonstration: gripper commands mark the closing and opening stages, and the commanded poses mark the path and the remaining endpoints.
The procedure uses task priors: the stage count and order, straight-line segments, and minimum descent and lift heights set from the cube size.
Its accuracy on held-out demonstrations is reported in Section~\ref{sec:robustness}.

\subsection{Executors and Stopping}
\label{sec:executors}
Both action representations use the same per-step servo.
It executes the first eight slots in order; at each step it converts the current slot's pose into a delta from the current end-effector pose and sends it to the end-effector delta-pose controller, which clips each axis to 0.1\,m and 0.1\,rad.
A dense slot holds the next step's target, so the command reproduces the demonstrated step and is not capped.
A keypose slot holds a stage endpoint, so the servo scales the translation and the rotation of the command separately to the 95th percentile of the per-step commands in the training demonstrations, keeping their directions; translation and rotation need not arrive together.
This cap is the only difference between the two default executors, and both candidate sets in Section~\ref{sec:selection} contain rules with and without it.
For the stopping comparison (Section~\ref{sec:stopping}), both policies also predict for every slot whether it lies past the end of the demonstrated task (a binary head with loss weight 0.1, thresholded at probability 0.5).
The predicted end is the first slot of the chunk's all-positive suffix; when it falls within the next eight slots, the policy executes the slots before it and then stops querying and commanding, the actuators keeping their last targets.

\subsection{Matched Training}
\emph{Policy architecture} denotes everything from the observation encoders to the action head.
Each action representation trains its own policy with the same policy architecture and output size (50 slots of pose and gripper); no weights are shared.
We use three policy architectures.
ACT is trained with batch 256 and 90k updates for 100 demonstrations, scaled with $\sqrt{N}$ for $N$ demonstrations, the same for both action representations.
Chunk length, execution rule and update count were selected for both action representations on tuning scenes from chunk lengths $\{10,30,50\}$, three execution rules (per-step, capped and temporally ensembled; Section~\ref{sec:selection}) and $\{30\mathrm{k},90\mathrm{k}\}$ updates; both chose $Q=50$, dense with per-step and keypose with capped execution.
ACT with a flow-matching head~\cite{lipman2023flow} (noisy action and time embeddings as decoder inputs, 10 Euler steps) reuses this configuration untuned.
$\pi_{0.5}$, a pretrained VLA policy, is fine-tuned from its released base policy with the RLinf~\cite{yu2025rlinf} implementation on the same chunks and batch order, with 7,500 updates selected on tuning scenes for both action representations; it is evaluated on StackCube only.
It runs without proprioceptive input, the RLinf default, because we found that with the discretized state in its prompt the dense $\pi_{0.5}$ policy stayed far below dense ACT on tuning scenes.

\subsection{Selection, Evaluation and Statistics}
\label{sec:selection}
Tuning scenes (60) serve only for selection.
Each comparison is evaluated on \MainScenes{} test scenes; selected execution is evaluated on \MainScenes{} fresh scenes that no selection used.
\emph{Default execution} runs dense slots as they are and keypose slots with the cap.
\emph{Selected execution} is chosen on tuning scenes with equal budgets from seven candidates per action representation, three of them shared: per-step, capped and ACT's temporally ensembled execution.
Dense adds four faster rules: stride $s\in\{1.5,2,3\}$ and stride 2 with the cap, where stride $s$ advances $s$ slots per control step with pose interpolation, slows to one slot per step around gripper changes, and queries again once it reaches the ninth slot (after six steps at stride 1.5); keypose adds four relaxed caps ($\times1.25$, $\times1.5$, $\times2$, $\times3$).
On clean tuning scenes dense selects stride 2 with the cap, and on level-3 tuning scenes (Section~\ref{sec:perturbations}) stride 1.5; keypose keeps its default in both.
Selected execution reuses the trained policies; at level~3 it is evaluated with four trained policies per action representation (Table~\ref{tab:main}), two of which were also used to choose the execution rule.

The primary metric is success within the time limit, that is, task success at any step of the episode; success held at the end of the episode is secondary.
Each task's limit is set once on 100 demonstrations from its native ManiSkill3 expert, as 1.5 times the longest one rounded up to 50 steps, and kept for the task's larger demonstration sets and demonstration variants so that results within a task are comparable.
The longest of these 100 demonstrations have 127, 246 and 226 steps, which gives 200 steps for StackCube, \HorizonStackPyramid{} for StackPyramid and \HorizonLiftPegUpright{} for LiftPegUpright.
On their own, StackPyramid's demonstrations from the revised expert (Section~\ref{sec:conditions}), at most 156 steps, would give 250, and LiftPegUpright's demonstrations re-collected with a slower executor, at most 222, the same \HorizonLiftPegUpright{}.
The perturbation levels keep the 200-step limit by this rule, although their demonstrations are longer.
We count execution speed as part of each action representation's performance, so a policy that slows down by reproducing the demonstrations' timing loses success within the limit; 400-step results, which separate slower completion from failure, are given alongside.

Each comparison reports the success difference $\Delta$, keypose minus dense in percentage points (pp), with a 95\% two-level bootstrap interval from 10,000 replicates, written $\Delta$ [lower, upper]; each replicate resamples the test scenes once, using the same resampled scenes for every success rate in the comparison, and resamples the training seeds independently for each action representation and setting.
An interval that excludes 0 marks a detectable difference, an interval within $\pm$5\,pp establishes equivalence, and any other interval is inconclusive; each comparison is judged on its own interval, without adjusting for the number of comparisons in the paper.
The main 100-demonstration comparison uses a two-stage design: 6 seeds per action representation, extended once to 12 if the interval includes 0 without establishing equivalence; it reached 12 seeds.

\subsection{Demonstration Perturbations and Scene Shift}
\label{sec:perturbations}
To vary how the demonstrations move without changing the task, each arm movement of the scripted expert (approach, descend, lift and place) is perturbed at four levels that jointly strengthen time scaling (by factors from 0.5 to 2 at level~3), a pause of 3--10 steps inside the movement (probability up to 0.6), detours through offset points (up to 8\,cm) and smooth joint noise (up to 0.02\,rad); detours and noise apply only to the approach and place movements, not to the short descend and lift around the grasp.
Only demonstrations that succeed and replay successfully are kept.
The mean demonstration length grows from 104.6 to 117.3, 135.2 and 167.1 steps, and the keypose speed cap is recomputed from each training set by the same rule.
\emph{Shifted conditions} change only the evaluation scenes: object scaling changes both cube edge lengths, with spacing and success tolerance following the scaled geometry, and position extrapolation places both cubes outside the training placement box ($|x|\le0.2$, $|y|\le0.3$\,m) but within $|x|\le0.25$, $|y|\le0.35$\,m and 0.8\,m of the robot base.
They use the execution selected on clean tuning scenes, with nothing reselected, and count only scenes the expert can solve: \MainScenesScaleLarge, \MainScenesScaleSmall{} and \MainScenesPosition{} test scenes under 1.15$\times$, 0.85$\times$ and position extrapolation, and \FollowUpScenesScaleLarge, \FreshScenesScaleSmall{} and \FreshScenesPosition{} fresh scenes.

\section{Experiments}
\label{sec:experiments}

The experiments address three questions: (Q1) which action representation succeeds more often with clean demonstrations, including against dense execution accelerated on tuning scenes (Section~\ref{sec:indist}); (Q2) which is more robust to demonstration variation, across endpoint sources and policy architectures, and to object-size and position shift (Sections~\ref{sec:robustness} and~\ref{sec:shift}); and (Q3) which conditions a keypose advantage depends on, what violating them costs, and how the two compare on a real robot (Sections~\ref{sec:stopping}--\ref{sec:real}).

\begin{table*}[t]
\centering
\caption{Keypose versus dense: success (\%) within 200 steps, 100-demonstration StackCube, ACT with planner endpoints and default execution unless stated. Selected execution: each action representation's execution rule chosen on clean in-distribution tuning scenes (level-3 tuning scenes for the level-3 rows), evaluated on fresh scenes; stride-1.5 dense: the best faster dense rule on the test scenes, evaluated on fresh scenes unless stated. Seeds per action representation (dense\,/\,keypose if they differ); $\Delta$ = keypose $-$ dense (pp) with 95\% intervals; bold: interval excludes 0; $^\dagger$: within the $\pm$5\,pp equivalence margin.}
\label{tab:main}
\footnotesize
\setlength{\tabcolsep}{5pt}
\begin{tabular}{@{}llcccl@{}}
\toprule
 & Setting & Seeds & Dense & Keypose & $\Delta$ [95\% interval] \\
\midrule
\multirow{3}{*}{Policy architecture} & ACT, L1 head & 12 & 79.7 & 83.6 & \textbf{\boldmath +3.9 [+0.6, +7.2]} \\
 & ACT, flow-matching head & 6 & 74.6 & 82.0 & \textbf{\boldmath +7.4 [+3.1, +11.8]} \\
 & stateless $\pi_{0.5}$ & 3 & 74.4 & 82.6 & \textbf{\boldmath +8.1 [+2.6, +13.6]} \\
\midrule
\multirow{3}{*}{Demonstrations} & 250 demos & 6 & 93.7 & 89.2 & \textbf{\boldmath $-$4.5 [$-$8.3, $-$1.1]} \\
 & 500 demos & 6 & 95.3 & 94.9 & $-$0.4 [$-$3.1, +2.2]$^\dagger$ \\
 & 1,000 demos & 6 & 97.3 & 96.4 & $-$0.8 [$-$2.4, +0.7]$^\dagger$ \\
\midrule
\multirow{2}{*}{\begin{tabular}[c]{@{}l@{}}Perturbation \\ level 3\end{tabular}} & 250 demos & 4 & 62.6 & 96.3 & \textbf{\boldmath +33.8 [+28.5, +39.3]} \\
 & ISR-resampled dense, execution tuned & 4 & 46.0 & 86.2 & \textbf{\boldmath +40.2 [+31.8, +48.1]} \\
\midrule
\multirow{1}{*}{Learned stop} & both action representations & 6 & 79.5 & 84.6 & \textbf{\boldmath +5.1 [+0.1, +10.2]} \\
\midrule
\multirow{3}{*}{Distribution shift} & object scale 1.15$\times$ & 12 & 67.7 & 80.7 & \textbf{\boldmath +13.0 [+8.4, +17.7]} \\
 & object scale 0.85$\times$ & 12 & 45.3 & 55.9 & \textbf{\boldmath +10.6 [+3.4, +17.8]} \\
 & position extrapolation & 12 & 21.2 & 31.5 & \textbf{\boldmath +10.4 [+5.3, +15.2]} \\
\midrule
\multirow{2}{*}{\begin{tabular}[c]{@{}l@{}}Selected execution, \\ in distribution\end{tabular}} & planner endpoints & 6 & 76.2 & 82.0 & \textbf{\boldmath +5.8 [+1.5, +10.1]} \\
 & extracted endpoints & 6\,/\,4 & 76.2 & 80.4 & +4.2 [$-$0.3, +8.7] \\
\midrule
\multirow{2}{*}{\begin{tabular}[c]{@{}l@{}}Selected execution, \\ perturbation level 3\end{tabular}} & planner endpoints & 4 & 63.7 & 84.7 & \textbf{\boldmath +20.9 [+14.9, +27.2]} \\
 & extracted endpoints & 4 & 63.7 & 84.5 & \textbf{\boldmath +20.8 [+14.7, +27.2]} \\
\midrule
\multirow{4}{*}{\begin{tabular}[c]{@{}l@{}}Selected execution, \\ distribution shift\end{tabular}} & object scale 1.15$\times$, planner endpoints & 6 & 72.3 & 80.1 & \textbf{\boldmath +7.8 [+1.4, +14.6]} \\
 & object scale 1.15$\times$, extracted endpoints & 6\,/\,4 & 72.3 & 80.5 & \textbf{\boldmath +8.2 [+1.3, +15.5]} \\
 & object scale 0.85$\times$, planner endpoints & 6 & 32.4 & 47.9 & \textbf{\boldmath +15.5 [+5.4, +25.8]} \\
 & position extrapolation, planner endpoints & 6 & 21.8 & 29.9 & \textbf{\boldmath +8.1 [+2.9, +13.3]} \\
\midrule
\multirow{5}{*}{\begin{tabular}[c]{@{}l@{}}Against \\ stride-1.5 dense\end{tabular}} & in distribution & 6 & 77.4 & 82.0 & \textbf{\boldmath +4.6 [+0.4, +8.9]} \\
 & object scale 1.15$\times$ & 6 & 69.9 & 80.1 & \textbf{\boldmath +10.2 [+3.6, +18.0]} \\
 & object scale 0.85$\times$ & 6 & 39.5 & 47.9 & +8.4 [$-$1.7, +17.9] \\
 & position extrapolation & 6 & 23.8 & 29.9 & +6.0 [$-$0.5, +12.1] \\
 & level 3, 250 demos, test scenes & 4 & 80.5 & 96.3 & \textbf{\boldmath +15.8 [+11.9, +20.1]} \\
\midrule
\multirow{3}{*}{Other tasks} & StackPyramid, 1,000 demos, 400 steps & 4 & 84.1 & 83.0 & $-$1.1 [$-$7.3, +5.3] \\
 & LiftPegUpright, 100 demos, 350 steps & 4 & 82.2 & 84.2 & +1.9 [$-$3.3, +7.8] \\
 & StackPyramid, 100 native demos, 400 steps & 4 & 8.5 & 13.2 & +4.8 [$-$1.1, +10.5] \\
\bottomrule
\end{tabular}
\end{table*}

\subsection{In Distribution and Demonstration Count}
\label{sec:indist}
With 100 clean demonstrations and default execution, keypose leads dense by $+$3.9\,pp, a detectable difference (Table~\ref{tab:main}).
With execution selected on tuning scenes (Section~\ref{sec:selection}), keypose's lead on fresh scenes is $+$5.8, also detectable.
The default- and selected-execution rows of Table~\ref{tab:main} differ in trained policies and scenes: default execution gives \DefaultExecDeltaID{} on the same fresh scenes, so the step from $+$3.9 to $+$5.8 is not a gain from execution selection.
The selected dense execution shortens the median number of steps to first success in successful episodes from 102 to 83 without a detectable gain in success ($+$0.6 [$-$3.8, $+$5.1]), while keypose needs 70; how much of keypose's lead comes from finishing sooner is not identified.
Against stride~1.5, the best faster dense rule when chosen on the test scenes, keypose also leads detectably on fresh scenes ($+$4.6, Table~\ref{tab:main}).
The lead recurs with the flow-matching head and with $\pi_{0.5}$ (Table~\ref{tab:main}).
With extracted endpoints, keypose against selected dense execution is $+$4.2, inconclusive; the keypose policies here use different training seeds from those with planner endpoints.

With more clean demonstrations and default execution, the in-distribution advantage does not persist: dense is detectably better at 250 demonstrations ($-$4.5), and at 500 and 1,000 demonstrations the two meet the $\pm$5\,pp equivalence rule.

\subsection{Robustness to Demonstration Variation}
\label{sec:robustness}
\begin{figure}[t]
  \centering
  \includegraphics[width=\columnwidth]{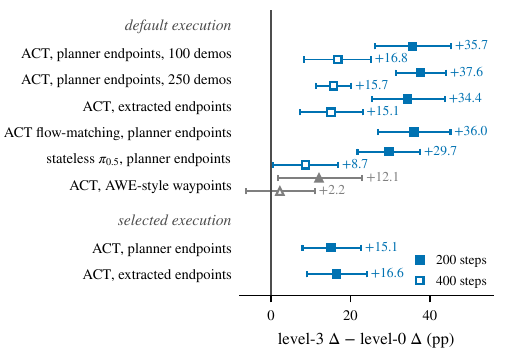}
  \caption{Robustness to demonstration variation in StackCube, 100 demonstrations per level unless stated:
  interaction of action representation and perturbation level (keypose $-$ dense at level~3 minus the same
  difference at level~0; filled within 200 steps, hollow within 400 steps where measured), with 95\% intervals. Top: default execution rules on
  300 test scenes, 4 training seeds per action representation and level (3 for $\pi_{0.5}$); the grey row is the same interaction for
  AWE-style waypoints instead of keypose.
  Bottom: dense and keypose use execution rules selected on tuning scenes, evaluated on 300 fresh scenes. Level-0
  dense\,/\,keypose seed counts are 6\,/\,6 for planner endpoints and 6\,/\,4 for extracted endpoints; at level~3
  both use the four trained policies of the top block.
  The rows share controls and are
  related evidence, not independent replications.}
  \label{fig:interactions}
\end{figure}

Perturbed demonstrations are longer, but success is still counted within 200 steps (Section~\ref{sec:selection}), with 400-step results alongside.
In Fig.~\ref{fig:hero}(b), dense success falls from 80.2\% at level~0 to 47.5\% at level~3, while keypose with planner endpoints stays between 78.5\% and 86.2\%.
Planner endpoints do not change with the perturbation level (Section~\ref{sec:representations}), so the endpoints extracted from the perturbed demonstrations below test whether the advantage holds when the endpoints themselves vary.
The interaction of action representation and perturbation level, $\Delta$ at level~3 minus $\Delta$ at level~0, is a detectable $+$35.7\,pp (Fig.~\ref{fig:interactions}), and $+$16.8 within 400 steps.
Among episodes successful within 400 steps, the median steps to first success grow from 103 to 172 for dense and stay at 70--75 for keypose; we read this as dense copying the demonstrations' speed changes and detours, an interpretation based on timing alone.
AWE-style waypoints keep a small part of the advantage (interaction $+$12.1, Fig.~\ref{fig:interactions}), inconclusive within 400 steps ($+$2.2).
They are extracted from each perturbed demonstration by AWE's dynamic program over positions, with gripper changes forced as waypoints and a 0.5\,mm threshold that gives about one waypoint per eight steps at level~0.
Endpoints extracted by the frozen procedure retain the measured advantage (interaction $+$34.4, Fig.~\ref{fig:interactions}).
On 100 newly collected held-out demonstrations at each of levels~2 and~3, 99.3\% (level~2) and 98.2\% (level~3) of extracted endpoints lie within 5\,mm and 2$^\circ$ of the expert's, measured against the nearer of the expert's commanded pose at the stage boundary and its first commanded pose of the next stage (87.0\% at level~3 against the boundary pose alone).
Extracted endpoints are not copies of planner endpoints: on the level-3 held-out demonstrations, 39\% of keypose slots differ from the planner-endpoint slots by more than 5\,mm.

With 250 demonstrations (levels~0 and~3 only), ACT using planner endpoints retains a detectable level-3 advantage of $+$33.8\,pp (Table~\ref{tab:main}) and an interaction of $+$37.6\,pp (Fig.~\ref{fig:interactions}).
Within 400 steps these remain positive at $+$12.2 [$+$8.9, $+$15.4] and $+$15.7; between 200 and 400 steps dense gains 269 of its 1,200 level-3 episodes and keypose 10.
The interaction difference from 100 demonstrations is unresolved: $+$1.9 [$-$8.9, $+$13.2].
Level~0 here reuses four of the six trained policies of Table~\ref{tab:main}'s 250-demonstration comparison.
The clean and perturbed 250-demonstration sets are drawn from largely overlapping scenes, but each keeps only the demonstrations that succeed and replay (Section~\ref{sec:perturbations}), so they are not pairs of the same scenes.
From 100 to 250 demonstrations three things also change together: the data, the update count (scaled with $\sqrt{N}$) and keypose's speed cap, which is recomputed from each training set (Section~\ref{sec:executors}).
Against stride~1.5, carried over from 100 demonstrations without reselection, keypose leads detectably by $+$15.8 at level~3.

\begin{table}[t]
\centering
\caption{Acceleration against a longer time limit at perturbation level~3: the same trained ACT policies (4 training seeds per action representation, planner endpoints) on 300 fresh scenes, counted within 200 and 400 steps; default and stride 1.5 are dense execution rules. Top: success (\%); middle: $\Delta$ (pp) at the same limit, the two dense executions resampled together; bottom: keypose within 200 steps against dense within 400. 95\% intervals, all of which exclude 0.}
\label{tab:horizon}
\footnotesize
\setlength{\tabcolsep}{4pt}
\begin{tabular}{@{}lcc@{}}
\toprule
 & 200 steps & 400 steps \\
\midrule
dense, default & 45.7 & 71.1 \\
dense, stride 1.5 & 63.7 & 75.6 \\
keypose & 84.7 & 88.8 \\
\midrule
stride 1.5 $-$ default & +18.1 [+14.2, +22.0] & +4.5 [+0.5, +8.5] \\
keypose $-$ default & +39.0 [+33.1, +45.2] & +17.8 [+11.9, +24.0] \\
keypose $-$ stride 1.5 & +20.9 [+14.9, +27.2] & +13.2 [+7.8, +18.6] \\
\midrule
\multicolumn{2}{@{}l}{keypose (200) $-$ default dense (400)} & +13.6 [+7.6, +19.8] \\
\multicolumn{2}{@{}l}{keypose (200) $-$ stride-1.5 dense (400)} & +9.1 [+3.4, +14.5] \\
\bottomrule
\end{tabular}
\end{table}

With planner endpoints the interaction also appears under the flow-matching head and $\pi_{0.5}$ ($+$36.0 and $+$29.7, Fig.~\ref{fig:interactions}), with $\pi_{0.5}$ reaching 52.2\% for dense and 90.0\% for keypose at level~3.
These replicate the direction of the effect; the flow and L1 interactions differ by $+$0.3 [$-$12.4, $+$13.1], too wide an interval to establish equal robustness.
With execution selected on level-3 tuning scenes, dense switches to stride~1.5, which raises its success on fresh scenes by $+$18.1 over its default execution (Table~\ref{tab:horizon}), and keypose still leads by $+$20.9 with planner endpoints and $+$20.8 with extracted endpoints (Table~\ref{tab:main}); default execution gives \DefaultExecDeltaLevelThree{} on the same fresh scenes.
The interaction under selected execution is $+$15.1 with planner endpoints and $+$16.6 with extracted endpoints (Fig.~\ref{fig:interactions}, bottom); because dense runs stride~2 with the cap at level~0 but stride~1.5 at level~3, it mixes the effect of the perturbation with the change of dense execution rule.
Accelerated dense remains slower (median 125 against 71 steps to first success).
Continuing these episodes to 400 steps recovers most, but not all, of the acceleration gain, and keypose within 200 steps stays ahead of both dense executions within 400 (Table~\ref{tab:horizon}).

Temporal standardization of the dense demonstrations does not close the gap.
An ISR-style resampling with the original paper's parameters and two adaptations to the delta-pose controller (gripper dwell frames kept, merges limited to what one control step can express) shows no detectable change from the original dense at level~3 ($-$1.5 [$-$10.9, $+$8.5]), and keypose leads it by $+$40.2 (Table~\ref{tab:main}).
It collapses to 24.2\% on clean demonstrations, so its interaction is $-$18.8 [$-$29.4, $-$8.5]; the negative interaction is driven by clean-data collapse and cannot support a robustness victory.
The other candidate execution rule and the 400-step limit give the same picture.
The physical consistency of the resampled transitions was not verified, and the cause of the collapse is not identified.

\subsection{Object-Size and Position Shift}
\label{sec:shift}
Under default execution, keypose leads with ACT under all three shifts, by $+$10.4 to $+$13.0\,pp (Table~\ref{tab:main}), and under 1.15$\times$ object scaling also with the flow-matching head ($+$12.9) and $\pi_{0.5}$ ($+$13.3).
With ACT, planner endpoints and both action representations using the execution selected on clean tuning scenes, keypose stays ahead on fresh scenes under all three shifts, by $+$7.8 to $+$15.5\,pp (Table~\ref{tab:main}); default execution gives \DefaultExecDeltaScaleLarge, \DefaultExecDeltaScaleSmall{} (inconclusive) and \DefaultExecDeltaPosition{} on the same fresh scenes.
Extracted endpoints keep the selected-execution advantage at 1.15$\times$ ($+$8.2); the keypose policies here use different training seeds from those with planner endpoints.
At 0.85$\times$, selected minus default dense success is \SelectedMinusDefaultDenseScaleSmall\,pp (inconclusive), so part of keypose's $+$15.5 lead there comes from the selected execution, and keypose's leads over default dense ($+$9.3 [$-$0.7, $+$19.1]) and stride-1.5 dense ($+$8.4, Table~\ref{tab:main}) are inconclusive.
Under position extrapolation, stride~1.5 helps dense ($+$6.4 [$+$1.0, $+$12.2]); keypose is numerically higher than stride-1.5 dense, inconclusive on expert-solvable scenes ($+$6.0, Table~\ref{tab:main}) and positive on all scenes ($+$6.5 [$+$0.8, $+$11.9]).
On the test scenes, with the three dense training seeds used to choose stride~1.5 (Section~\ref{sec:indist}), keypose was 2.6\,pp below stride-1.5 dense (inconclusive); the two measurements differ in both scenes and training seeds.
Only planner endpoints were evaluated under 0.85$\times$ and position extrapolation.
Under selected execution, a larger advantage out of distribution than in distribution is not established ($+$7.8 at 1.15$\times$ against $+$5.8).

\begin{table}[t]
\centering
\caption{What keypose requires: success (\%) as the demonstrations, the execution rule or stopping change, with dense on the same demonstrations. Success within 350 steps for LiftPegUpright and 400 for StackPyramid; for StackCube, success still held at the end of the 200-step episode, which is what stopping affects. 4 training seeds (6 for StackCube).}
\label{tab:conditions}
\footnotesize
\setlength{\tabcolsep}{4pt}
\begin{tabular}{@{}llcc@{}}
\toprule
 & Setting & Dense & Keypose \\
\midrule
\multirow{2}{*}{\begin{tabular}[c]{@{}l@{}}StackCube, \\ 100 demos\end{tabular}} & no learned stop & 71.2 & 61.1 \\
 & learned stop for both & 77.4 & 77.2 \\
\midrule
\multirow{4}{*}{\begin{tabular}[c]{@{}l@{}}LiftPegUpright, \\ 100 demos\end{tabular}} & expert demos & 76.8 & 1.0 \\
 & expert demos, keypose synchronized & 76.8 & 24.0 \\
 & executor-collected demos & 72.6 & 67.6 \\
 & slower-executor demos & 85.1 & 86.6 \\
\midrule
\multirow{3}{*}{\begin{tabular}[c]{@{}l@{}}StackPyramid, \\ 1,000 demos\end{tabular}} & native expert demos & 69.4 & 44.9 \\
 & revised expert, plain servo & 82.4 & 58.1 \\
 & revised expert, reference servo & 82.4 & 82.1 \\
\bottomrule
\end{tabular}
\end{table}

\subsection{Learned Stopping}
\label{sec:stopping}
Without a learned stop, keypose holds success to the end of the episode less often than dense (Table~\ref{tab:conditions}), mainly because it more often loses success after reaching it (27.6\% of successful episodes against 10.3\%).
The same learned stop raises success held at the end of the episode for both to a similar level ($-$0.2 [$-$5.3, $+$5.1], no detectable difference), while keypose keeps a detectable lead in success within 200 steps ($+$5.1, Table~\ref{tab:main}).

\subsection{Executor Compatibility and Further Tasks}
\label{sec:conditions}
Between endpoints, keypose's motion is set by its executor (Section~\ref{sec:executors}), so the demonstrations and the execution rule must be compatible with that motion.
Table~\ref{tab:conditions} shows two kinds of incompatibility on \ConditionsOtherScenes{} scenes per task, with dense, trained on the same demonstrations and kept at its default execution, as a reference.
First, the executor's path can conflict with the task physics.
On LiftPegUpright, which lifts a peg lying on the table to an upright pose, with 100 expert demonstrations keypose predicts the right endpoints (median yaw error 2$^\circ$), but its executor completes the translation before the wrist finishes rotating.
The gripper then descends misaligned and jams on the peg: keypose reaches 1.0\% where dense reaches 76.8\%.
Synchronized execution, which paces translation and rotation to arrive together, recovers keypose only in part; more demonstrations of the same expert cannot supply the jammed states, which never occur in them.

Second, the executor may take the policy out of distribution: its motion between endpoints is not constrained by the demonstrations, so when it moves differently from them, the policy can be queried in states they do not cover, whereas dense, whose slots copy the demonstrated motion, stays near them.
Demonstrations collected by replaying the expert's endpoints through the keypose executor, or re-collected with a slower executor whose cap is halved, bring keypose close to or level with dense.
The execution rule must then match collection: switching keypose on executor-collected demonstrations to synchronized execution drops its tuning-scene success from \SyncSwapTuneBefore{} to \SyncSwapTuneAfter{} of 60 (mean of four seeds).
StackPyramid places a cube beside another, pushing it closer first if needed, and stacks a third on both; with 1,000 native expert demonstrations, keypose stays well below dense.
A revised expert replans the remaining endpoints from the executed state and collects its demonstrations with a reference servo, in which the keypose executor follows a reference pose that advances at most one capped step per control step; keypose then comes close to dense with the reference servo, which matches collection, but reaches only 58.1\% with the plain servo.

For each further task, the executor-compatible demonstrations and execution rule (the last rows of Table~\ref{tab:conditions}) were chosen among several variants on the task's tuning scenes and then frozen, so Table~\ref{tab:main} evaluates them on fresh scenes.
There the two action representations are close, $-$1.1 on StackPyramid at \HorizonStackPyramid{} steps and $+$1.9 on LiftPegUpright at \HorizonLiftPegUpright{} steps, without established equivalence, as they were on the tuning scenes ($-$0.2 and $+$1.5).
Within 200 steps, which its revised demonstrations fit, StackPyramid gives \StackPyramidTwoHundredDelta{} \StackPyramidTwoHundredCI{}.
With only 100 StackPyramid demonstrations both action representations stay below 14\% within 400 steps ($+$4.8, inconclusive; Table~\ref{tab:main}).

\subsection{Real-Robot Closed-Loop Comparison}
\label{sec:real}
We trained one ACT policy per action representation on human demonstrations of opening a drawer (105 collected, 90 for training and 15 held out), recorded with a hand-held gripper in the style of UMI~\cite{chi2024umi} and observed through its wrist camera.
Endpoints are extracted from gripper-width events and the poses reached at the end of four stages (approach, close, pull and release), and both action representations express poses relative to the current gripper frame.
To grasp the horizontal handle, the gripper turns by about 90$^\circ$ during the approach.
Each of \RealTrials{} initial conditions offsets the start pose of a 7-degree-of-freedom arm by up to 4\,cm sideways, 5\,cm toward or 1.5\,cm away from the handle and 5$^\circ$ in yaw; a trial succeeds once the drawer is 10\,cm open within \RealLimit\,s.
With default execution and no tuning trials, dense succeeded in \RealDenseSuccess{} of \RealTrials{} trials and keypose in \RealKeyposeDefaultSuccess{}.
Keypose's executor, moving toward each endpoint at the capped speeds, finished the turn well before reaching the handle, whereas the demonstrations do most of it in the last few centimeters.
After these failures, we set its rotation and forward speed from statistics of the training demonstrations to bring its turning and advance as close to theirs as possible, which raised its success to \RealKeyposeAdjustedSuccess{}.
As only the execution changed, these failures appear to be the second incompatibility of Section~\ref{sec:conditions}: execution that departs from the demonstrations, such as the early turn, takes keypose out of the states they cover.

\section{Conclusion}
\label{sec:conclusion}

In our controlled comparison in simulation, chunks written as stage endpoints were far more robust to demonstration variation, and stayed ahead of selected accelerated dense execution under object scaling and position extrapolation.
With clean demonstrations, keypose led only when they were scarce; with more it was level or behind, with extracted endpoints inconclusive, and on two further tasks close to dense, but only with demonstrations and execution rules suited to its executor.
When learning from limited human demonstrations on a real robot, dense was more reliable and keypose again depended on its executor: making only its execution follow the demonstrations recovered much of its gap to dense.
These results suggest that keypose trades sensitivity to demonstration variation for sensitivity to execution that leaves the states they cover.

The comparison is scoped to stage-structured manipulation, whose tasks can be divided into stages with definable endpoints.
The simulated demonstrations come from scripted experts and the demonstration variation is constructed, so robustness to the irregularities of human demonstrations was not isolated; extracted endpoints also rely on task priors.
The real-robot evidence covers one task and one trained policy per action representation, with keypose's execution adjusted after its failures.
Keypose also needs a learned stop.
Future work should evaluate keypose on more tasks, robot embodiments and policy architectures, with larger simulated and real datasets and with executors learned from the demonstrations.
It should also test keypose at scale, by training multi-task policies on mixed data from many tasks and by pretraining on large datasets in the keypose action representation.
Code and data will be released upon acceptance.


\bibliographystyle{IEEEtran}
\bibliography{references}

\end{document}